\documentclass[pmlr]{jmlr}
\usepackage{booktabs}
\usepackage[load-configurations=version-1]{siunitx} 

\jmlrvolume{1}
\jmlryear{2020}
\jmlrworkshop{NeurIPS2019 Competition \& Demonstration Track} 

\title[Retrospective Analysis of the 2019 MineRL Competition]{Retrospective Analysis of the 2019 MineRL Competition on Sample Efficient Reinforcement Learning}

 \author{\Name{Stephanie Milani} \Email{smilani@cs.cmu.edu}\\
 \addr{Machine Learning Department, Carnegie Mellon University}
 \AND
  \Name{Nicholay Topin} \Email{ntopin@cs.cmu.edu}\\
  \addr Machine Learning Department, Carnegie Mellon University
  \AND
  \Name{Brandon Houghton\nametag{\thanks{Work done as a research assistant at Carnegie Mellon University.}}} \Email{bhoughton@openai.com}\\
  \addr{OpenAI}
  \AND 
  \Name{William H. Guss} \Email{wguss@cs.cmu.edu}\\
  \addr OpenAI and Machine Learning Department, Carnegie Mellon University
  \AND
  \Name{Sharada P. Mohanty} \Email{mohanty@aicrowd.com}\\
  \addr AIcrowd
  \AND
  \Name{Keisuke Nakata} \Email{nakata@preferred.jp} \\
  \addr Preferred Networks
  \AND
  \Name{Oriol Vinyals} \Email{vinyals@google.com}\\
  \addr DeepMind
  \AND
  \Name{Noboru Sean Kuno} \Email{nkuno@microsoft.com}\\
  \addr Microsoft Research
  }

\editors{Hugo Jair Escalante and Raia Hadsell}

\usepackage{graphicx}
\usepackage{xcolor}

\begin{document}
\maketitle

\begin{abstract}
To facilitate research in the direction of sample efficient reinforcement learning, we held the MineRL Competition on Sample Efficient Reinforcement Learning Using Human Priors at the Thirty-third Conference on Neural Information Processing Systems (NeurIPS 2019).
The primary goal of this competition was to promote the development of algorithms that use human demonstrations alongside reinforcement learning to reduce the number of samples needed to solve complex, hierarchical, and sparse environments.
We describe the competition, outlining the primary challenge, the competition design, and the resources that we provided to the participants.
We provide an overview of the top solutions, each of which use deep reinforcement learning and/or imitation learning.
We also discuss the impact of our organizational decisions on the competition and future directions for improvement.
\end{abstract}
\begin{keywords}
reinforcement learning competition, reinforcement learning, imitation learning
\end{keywords}

\section{Introduction}
\label{sec:intro}
Many of the recent, celebrated successes of artificial intelligence (AI), such as AlphaStar~\citep{starcraft2019}, AlphaZero~\citep{alphazero}, and OpenAI Five\footnote{\url{https://openai.com/five}} achieve human- or super-human-level performance using deep reinforcement learning (DRL).
Thus far, improvements to the state of the art have used ever-increasing computational power\footnote{\url{https://blog.openai.com/ai-and-compute}}.
In part, this computational increase is due to the computation required per environment sample; however, it is primarily due to the increasing number of environment samples required for training these learning algorithms.
These growing computational requirements mean that a shrinking portion of the AI community can reproduce these results, let alone improve upon these systems.
One well-known way to reduce environment-sample complexity, and, by extension, make methods more accessible, is to leverage human demonstrations and priors over desired behavior~\citep{pfeiffer2018reinforced,dubey_humanpriors2018,zhang2019leveraging}.

To further the development of novel, sample efficient methods that leverage human priors for sequential decision-making problems,
we held the first-ever MineRL Competition on Sample Efficient Reinforcement Learning Using Human Priors\footnote{\url{http://minerl.io}} at the Thirty-third Conference on Neural Information Processing Systems (NeurIPS 2019)~\citep{guss_minerl_neurips2019}.
Teams developed procedures for training an agent to obtain a diamond in Minecraft.
To limit computational requirements, the learning procedures were evaluated by retraining them from random weights under a strict computational and environment-sample budget.
To assist participants with developing their algorithms, teams were provided with the largest-ever data set of human trajectories in Minecraft~\citep{guss_minerl_ijcai2019}.
Our challenge attracted over 1000 registrated participants with a total of 662 submissions\footnote{\url{https://www.aicrowd.com/challenges/neurips-2019-minerl-competition}}. 

In this work, we describe the competition paradigm and procedure, and provide a summary of the top nine solutions.
We identify common high-level approaches --- such as leveraging human demonstrations and using hierarchical reinforcement learning --- used in many of the top solutions.
We conclude by discussing the outcomes of our choices as organizers of the competition, including how we evaluated the submissions and specified the rules.

\section{Background}
\paragraph{Minecraft.}
Minecraft is a 3D, first-person, open-world game revolving around gathering resources and creating structures and items.
Because it is a sandbox environment, designers can devise a variety of goals and challenges for intelligent agents.
Additionally, because it is an embodied domain and the agent's surroundings are varied and dynamic, it presents many of the same challenges found in real-world control tasks, such as planning over long time horizons and determining a good representation of the environment~\citep{alterovitz2016robot}.
As a result, it is a promising and popular testbed for both single- and multi-agent reinforcement learning (RL) and planning algorithms~\citep{abel2015goalbased,aluru2015minecraft,abel2016gradboost,oh2016minecraft,tessler2017deephrl,shu2018multitask,Frazier2019ImprovingDR,master2019string,Arumugam2019drl,trott2019distance}.

\paragraph{Reinforcement Learning Competitions,}
To our knowledge, no previous competitions explicitly focused on using imitation learning (and, more generally, learning from demonstrations) alongside RL.
Most previous RL competitions~\citep{kidzinski_learn2runsynth2018,nichol_learnfast2018,perezliebana_marlo2019,juliani2019} focused on performing well on a given domain, and not on developing robust algorithms that can perform well on a broad set of domains.
Consequently, winning submissions often required hand-engineered features and stemmed from using large amounts of computational resources for training and development.

\section{Competition Overview}
We provide an overview of our competition.
We describe the primary task and environment in Section~\ref{sec:task}.
In Section~\ref{sec:comp-design}, we describe the structure of our competition.
In Section~\ref{sec:resources}, we describe the resources that we provided participants.

\subsection{Task}
\label{sec:task}
Competitors were tasked with solving the \texttt{ObtainDiamond} environment.
Solving the environment consists of controlling an embodied agent to obtain a diamond by navigating the complex item hierarchy of Minecraft.
In solving this task, a learning algorithm has direct access to a 64x64 pixel observation from the perspective of the embodied Minecraft agent, and a set of discrete observations consisting of each item required for obtaining a diamond that the agent has in its possession.
The action space is the Cartesian product of continuous view adjustment (turning and pitching), binary movement commands (left/right, forward/backward), and discrete actions for placing blocks, crafting items, smelting items, and mining/hitting enemies.

An agent receives reward for completing the full task of obtaining a diamond.
The full task of obtaining a diamond can be decomposed into a sequence of prerequisite subtasks of increasing difficulty. 
An agent also receives reward for the first time it accomplishes each subtask in the sequence.
An agent receives twice the reward as received for accomplishing the previous subtask (starting from a reward of $1$).
The exception to this rule is achieving the full \texttt{ObtainDiamond} task by obtaining a diamond: accomplishing the final task is worth four times as much as completing the previous subtask.

\begin{figure}[tbp]
   \floatconts
     {fig:test_environments}

     {%
       \subfigure[Original texture]{\label{fig:original}%
         \includegraphics[width=0.25\linewidth]{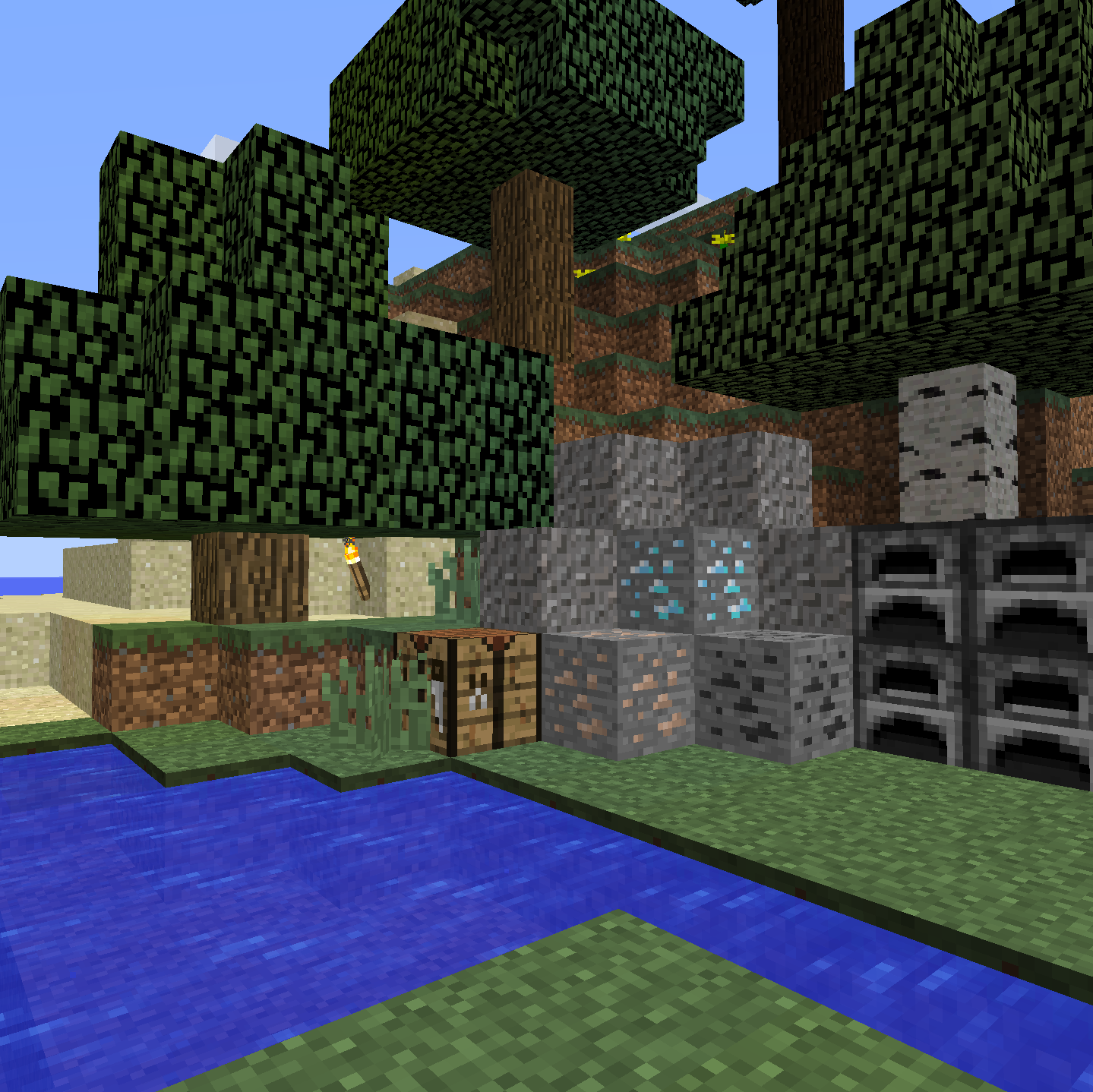}}%
       \qquad
       \subfigure[Validation texture]{\label{fig:validation}%
         \includegraphics[width=0.25\linewidth]{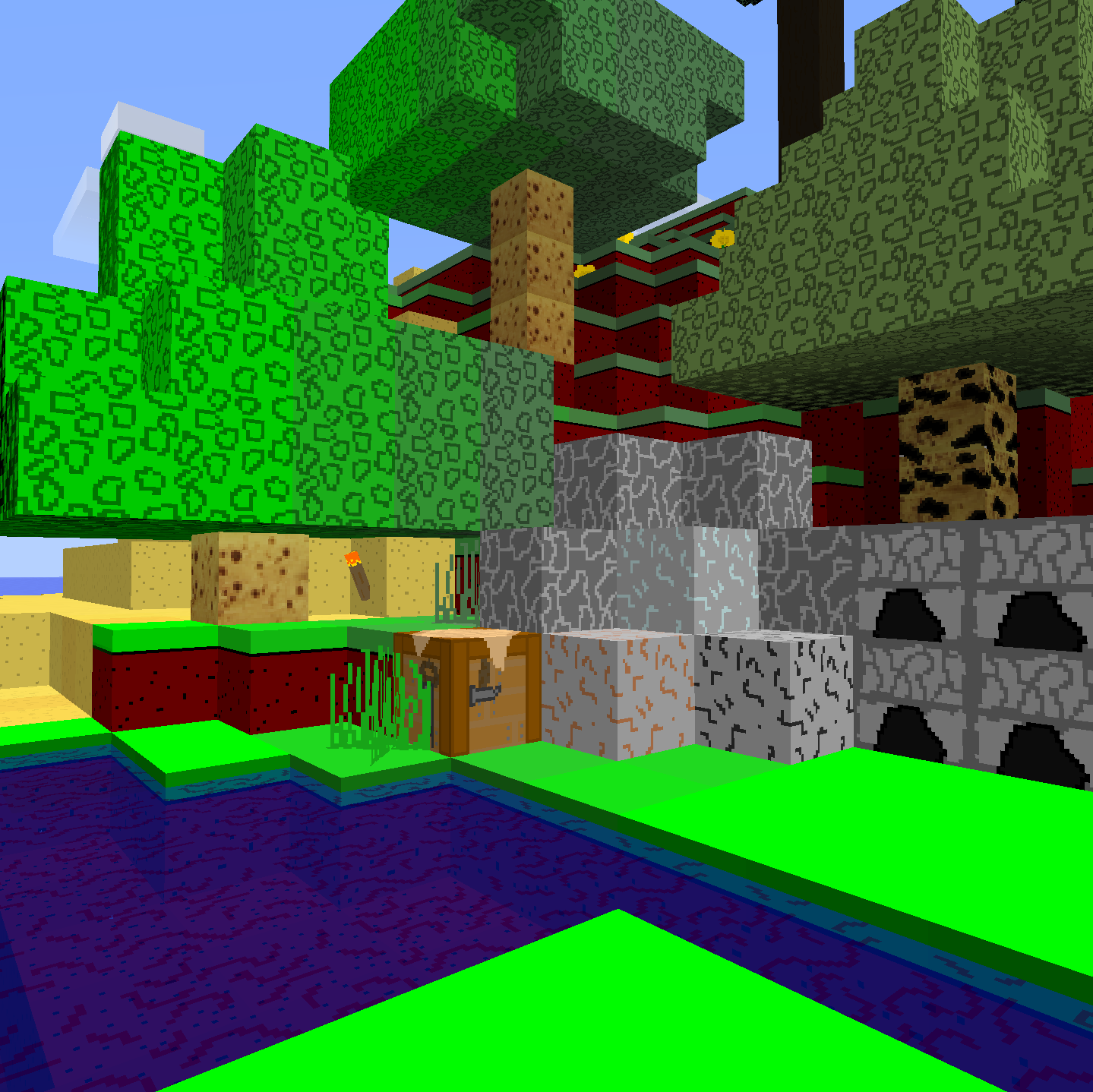}}
       \qquad
       \subfigure[Test texture]{\label{fig:test}%
         \includegraphics[width=0.25\linewidth]{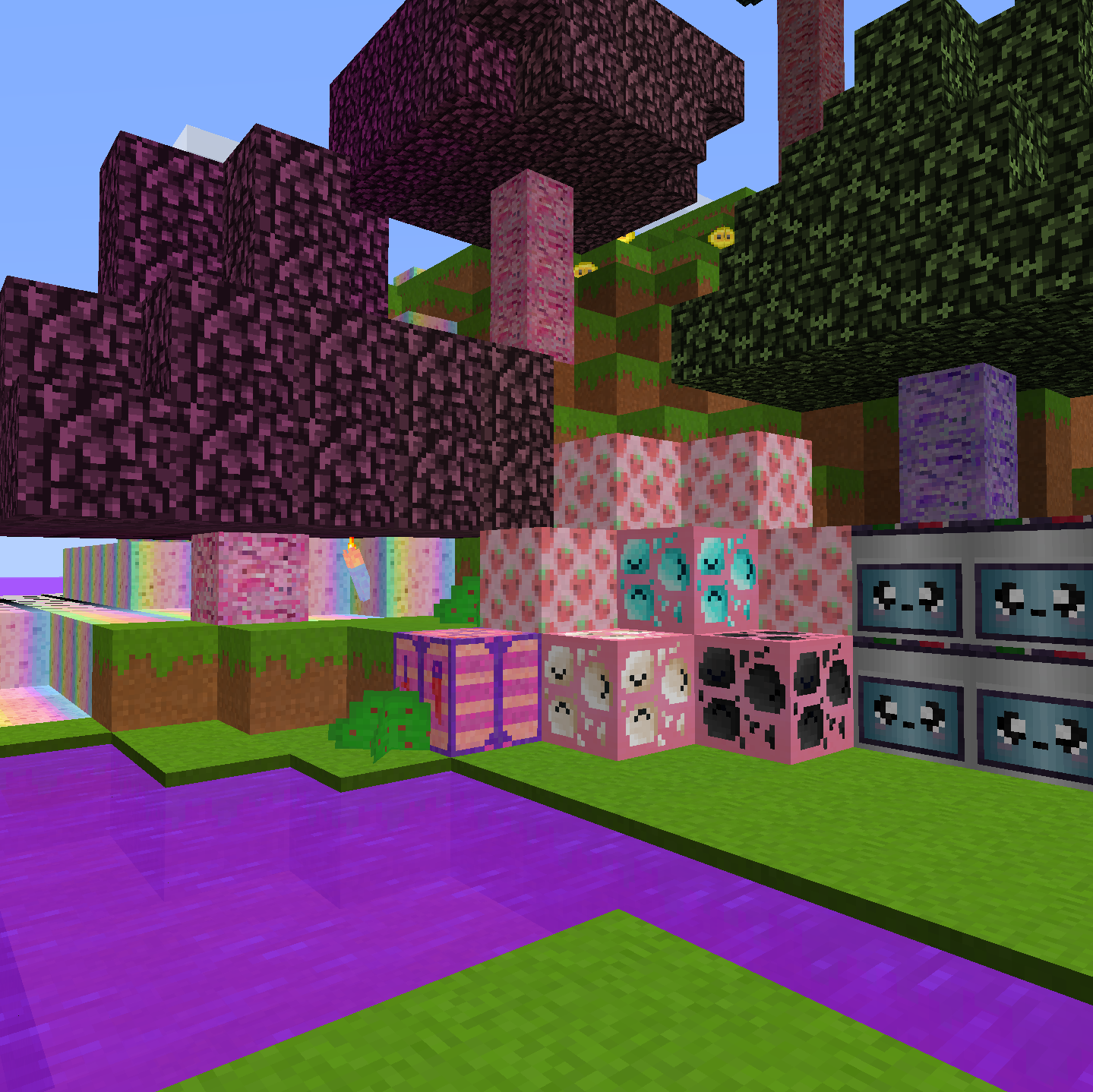}} 
     }

    \vspace{-0.6cm}
    {\caption{An example game state rendered with three different textures.}{\label{test_environments}}}
    \vspace{-0.2cm}
   \end{figure}

\subsection{Competition Design}
\label{sec:comp-design}

The competition was split into two rounds.
In Round 1, teams of up to six total participants trained their agents and submitted trained models for evaluation to determine their leaderboard ranks.
In this round, we limited each team to 25 total submissions.
At the end of Round 1, the teams submitted their source code.
Each submission received a score consisting of the average reward achieved by that submission over 100 episodes.
The top submission for each team was then reviewed by the rules committee to verify that they complied with the rules.
The 13 teams with the highest-scoring, rule-compliant submissions advanced to Round 2.
We invited these teams to
present their work at the NeurIPS 2019 competition workshop. 
With the support of Microsoft Research, we provided each team with a \$1200 USD travel grant to attend the conference.

In Round 2, each team could submit up to five learning procedures to be trained from random weights on a held-out test environment.
We independently trained each submission for six days using Azure NC6 and NC12 instances with a strict environment step budget of 8 million samples.
To encourage generalizability and compliance with rules, we retrained the solutions submitted in the final round using: 1) a perturbed action space, and 2) an entirely new, previously-unseen texture pack\footnote{The test texture pack we used can be found \href{https://www.planetminecraft.com/texture_pack/kawaii-world-3365736}{here}.} (illustrated in Figure~\ref{fig:test}).
We provided teams with a validation texture pack\footnote{We used a modified version of \href{https://www.planetminecraft.com/texture_pack/ragdoll-adventure}{this} texture pack as our validation texture pack.} (illustrated in Figure~\ref{fig:validation}) to use during training so that they could check whether their appeoaches could generalize to previously-unseen environments before submitting their learning procedures.
The final ranking was determined based on each team's highest-scoring submission.

Given the unique paradigm of training learning procedures, we set strict rules to encourage sample efficiency and generalizability, and prevent participants from loading pretrained models. 
Sample efficiency constraints were enforced by setting a strict environment sample limit of 8 million frames (or roughly 110 hours of in-game interactions), as well as limiting computation to 144 hours of wall clock time. 
We permitted frame skipping; however, each frame skipped counted against the sample budget. 
Participants were also free to shape the observations of the agent, but they could not directly encode them in the policy.
For example, switching sub-policies based on the number of logs collected was not permitted.
To avoid participants including pretrained models in their submissions, in Round~2, any files in the submitted code repositories that exceeded 4MB were programmatically deleted. 
A more detailed discussion of the competition paradigm can be found in previous work~\citep{houghton2019reproducibility}.

In both rounds, participants submitted their code as standalone repositories compatible with AIcrowd-repo2docker\footnote{\url{https://github.com/AIcrowd/repo2docker}}. These repositories included the code, trained models, and runtime specifications, which enabled us to build consistent Docker images out of the submissions. These images were orchestrated on a custom Kubernetes cluster while respecting any round-specific policies. 
We evaluated the submissions in complete network isolation to avoid any leak of information due to malicious code submitted by participants. 

\subsection{Resources}
\label{sec:resources}
In addition to the \texttt{ObtainDiamond} task, we created a number of auxiliary tasks, which we believed would be useful for solving \texttt{ObtainDiamond}.
These tasks were provided to participants as Gym environments~\citep{gym}.
Additionally, we provided the participants with a comprehensive starter pack\footnote{\url{https://github.com/minerllabs/competition_submission_starter_template}}, consisting of extensive documentation, starter code, a description of the submission procedure, a Microsoft Azure quick-start template, and the Docker images that we used to validate the training procedures during retraining.
We gave the participants a large dataset of human demonstrations~\citep{guss_minerl_ijcai2019}, which participants could use to train their agents.
Preferred Networks\footnote{\url{https://preferred.jp/en}} provided extensive baselines\footnote{\url{https://github.com/minerllabs/baselines}} implemented in ChainerRL~\citep{fujita2019chainerrl}.
Participants were able to incorporate these baselines into their submissions.
These baselines included behavioral cloning, deep Q-learning from demonstrations (DQfD)~\citep{dqfd}, Rainbow~\citep{rainbow}, generative adversarial inverse RL (GAIL)~\citep{gail_2016}, and proximal policy optimization (PPO)~\citep{ppo}. 

AIcrowd\footnote{\url{https://www.aicrowd.com}} provided a unified interface\footnote{\url{https://discourse.aicrowd.com/c/neurips-2019-minerl-competition}} for participants to sign-up for the competition, ask the organizers and other participants questions, and monitor their progress. 
We also created a public Discord server for more informal communication with participants. 

Through our generous sponsor, Microsoft Azure, we provided compute grants of \$500USD each in Micrsoft Azure\footnote{\url{https://azure.microsoft.com/en-us}} credits for 50 individuals that self identified as lacking access to the necessary compute power to participate in the competition. 
Additionally, thanks to Microsoft Azure, we provided each of the top teams from Round~1 with \$1500USD of Azure credits for their experiments in Round~2.

\begin{table}[tbp]
   \centering
   \begin{tabular}{ll}
   \begin{tabular}{|c|c|}
      \hline
      Team Name & Round 1 Score \\
      \hline 
      MeatMountain & 48.42 \\
      \hline
      xt & 46.75 \\
      \hline
      shadowyzy & 37.82 \\
      \hline 
      CraftRL & 33.15 \\
      \hline 
      CDS & 29.62 \\
      \hline 
      mc\_rl & 27.43 \\
      \hline
      I4DS & 23.96 \\
      \hline
      UEFDRL & 21.70 \\
      \hline 
      TD240 & 17.12 \\
      \hline 
   \end{tabular}
   &
   \begin{tabular}{|c|c|}
      \hline
      Team Name & Round 2 Score \\
      \hline 
      CDS & 61.61 \\
      \hline
      mc\_rl & 42.41 \\
      \hline
      i4DS & 40.80 \\
      \hline 
      CraftRL & 23.81 \\
      \hline 
      UEFDRL & 17.90 \\
      \hline 
      TD240 & 15.19 \\
      \hline
      LAIR & 14.73 \\
      \hline
      Elytra & 8.25 \\
      \hline 
      karolisram & 7.87 \\
      \hline 
   \end{tabular}
   \end{tabular}
   \caption{Scores of best-performing submissions of top teams in Round 1 and Round 2.}
   \label{table:finalleaderboard}
   \vspace{-0.2cm}
   \end{table}

\section{Outcomes for Participants}
We quantitatively and qualitatively analyze the submissions made by participants of our competition.
In Section~\ref{subsec:performance_overview}, we provide quantitative information about the submissions.
In Section~\ref{subsec:summary_nine}, we summarize the approaches used by the top nine teams in the final round.
In Section~\ref{subsec:special_awards}, we describe the special awards awarded as part of the competition.

\subsection{Submission Performance Overview} \label{subsec:performance_overview}
\begin{figure}[t]
  \floatconts
    {fig:itemscore}

    {\includegraphics[width=0.65\linewidth]{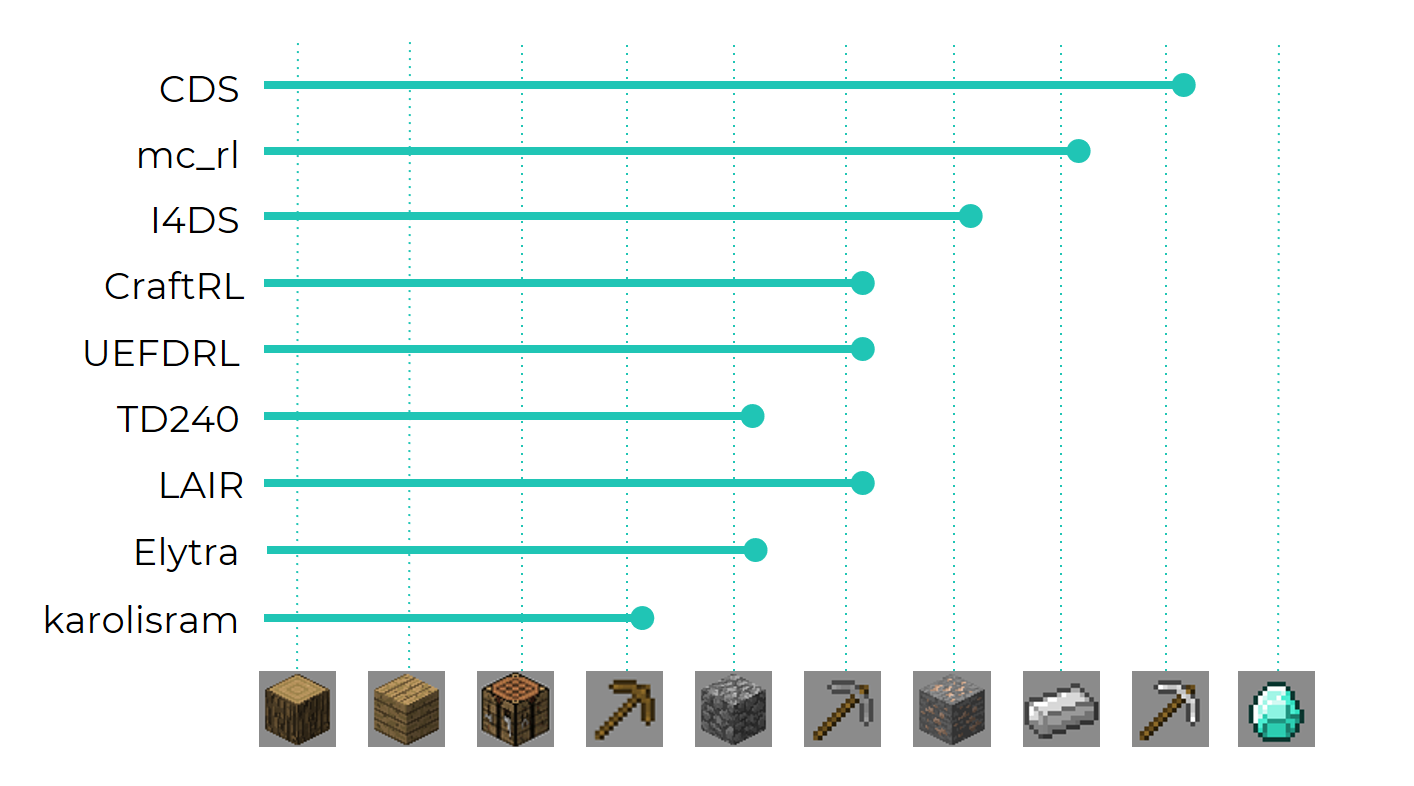}}

    \vspace{-0.8cm}
    {\caption{Maximum item score for each team over the 100 evaluation episodes in Round 2.}{\label{itemscore}}}
  \end{figure}

In Round 1, there were over 540 submissions, and 25 teams drastically outperformed the provided baselines.
Table~\ref{table:finalleaderboard} shows that the top 3 teams from Round 2 on average outperformed the top 3 teams from Round 1 (means: 48.27 and 44.33, respectively), despite the Round 2 agents being trained on a hold-out test environment. 
Perhaps unsurprisingly, the Round 1 scores had a higher mean (31.77) and lower standard deviation (10.22) than the Round 2 scores (25.84 and 17.37, respectively).
Additionally, the Round 1 scores had a smaller range (31.30) than the Round 2 scores (53.74).
Figure~\ref{itemscore} shows the maximum item score over the evaluation episodes of Round 2.
Although no team obtained a diamond, the top team obtained the penultimate prerequisite item to obtaining a diamond.

\subsection{Summary of Top Nine Solutions} \label{subsec:summary_nine}
The top four teams used some form of hierarchical RL, taking advantage of the hierarchicality of Minecraft. 
All teams used some form of action reduction (e.g., removing actions that were almost never performed by human experts) to manage the complexity of the environment.
Most teams leveraged the human data to improve the sample efficiency of their algorithms.

The top team, CDS, used a hierarchical deep Q-network with forgetting that uses an adaptive ratio for sampling expert demonstrations from a separate demonstration replay buffer\footnote{A video example of CDS's trained agent can be found \href{https://youtu.be/GHo8B4JMC38}{here}.}~\citep{skrynnik2019hierarchical}.
The second-place team, mc\_rl, trained their hierarchical policies entirely from human demonstrations with \textit{no} environment interactions\footnote{A video example of mc\_rl's trained agent can be found \href{https://youtu.be/W9-7FX4YZbI}{here}.}.
The third-place team, i4DS, used demonstrations to bootstrap their hierarchical RL algorithm, effectively using the human data to pretrain the models to predict recorded human actions from observations, then used RL to refine the resulting agents~\citep{scheller2020sample}.

The fourth-place team, CraftRL, aggregated grounded actions into options~\citep{options} and then used a meta-controller to select between options. 
They pretrained the meta-controller on demonstration data using behavioral cloning.
The fifth-place team, UEFDRL, used a single deep residual neural network trained to mimic human actions from the MineRL dataset~\citep{kanervisto2020playing}. This team used action discretization to deal with the complexity of the environment.
The sixth-place team, TD240, used a discriminator soft actor critic, which uses a discriminator based on adversarial inverse RL as a reward function.

The seventh-place team, LAIR, used meta-learning shared hierarchies (MLSH)~\citep{mlsh}, a hierarchical RL algorithm, and pretrained the master and subpolicies with human data before training MLSH with environment interactions.
The eighth-place team, Elytra submitted a Rainbow baseline with modifications, including adding a small amount of stochasticity to the environment and restricting the camera movement to 1 degree of freedom.
However, they explored the idea of training an ensemble of value functions and believe this approach to be promising.
The ninth-place team, karolisram, used the PPO baseline with modifications, including removing frame-skip and reducing the action space.

\begin{table}[tbp]
   \centering
   \begin{tabular}{|c|c|c|c|}
      \hline
      & Use hierarchical      & Use human & Use action \\
      & RL & data      & reduction \\
      \hline
      Number of teams & 6 & 7 & 9 \\
      \hline
      Percent of teams & 67\% & 78\% & 100\% \\
      \hline 
   \end{tabular}
   \caption{Overview of general approaches taken by the top 9 teams.}
   \vspace{-0.2cm}
   \label{table:overview-approaches}
   \end{table}

\subsection{Special Awards} \label{subsec:special_awards}
In addition to evaluating teams on the performance of their learning algorithms, we awarded teams prizes for research contributions. 
Since this was the first year in which the competition was held, we awarded two prizes to extreme and opposite research paradigms: one based purely on imitation learning and the other one based purely on RL, with no use of human priors. 
We awarded the main research prize to mc\_rl for the former approach and the runner-up research prize to karolisram for the latter approach.
We hope that these special awards will encourage future participants to maximally explore the space of solutions, even if this comes at the expense of performance.

\section{Organizational Outcomes}
We describe important organizational outcomes of the competition.
We begin by presenting the impact of our chosen rule set in Section~\ref{subsec:rule_spec}.
Then, in Section~\ref{subsec:provided_package}, we discuss the effectiveness of our distribution of competition materials.
Finally, we summarize the effects of our chosen evaluation methodology in Section~\ref{subsec:eval_results}.

\subsection{Repercussions of the Rules} \label{subsec:rule_spec}
Unlike many other DRL competitions, we sought to limit the use of domain knowledge through feature engineering and hard-coded policies.
Though this ruleset has led to the use of more general methods, it also led to a lot of clarifying questions from participants.
Since these were asked across various communication channels
, there were duplicate questions and no single, comprehensive answer.
In future renditions of this competition, we will have a clear place for rule clarifications and a more robust restriction on use of domain knowledge.

A secondary effect of the limitation on domain knowledge use was the difficulty in checking Round 1 submissions for this rule violation. This rule was the only one which led to complications of this type; restrictions on training time and number of samples are quantitative restrictions and therefore easy to enforce. Since competitor's submissions were not trained on a ``hold out'' environment, a committee spent more work hours than expected manually reviewing the code to identify manually specified policies and other violations.  

\subsection{Effectiveness of Competition Materials} \label{subsec:provided_package}
As mentioned in Section~\ref{sec:resources}, we provided participants with the tasks as Gym environments, the dataset, and baseline method implementations.
The environments and dataset were both complete before the competition began, but the baselines were developed after the competition was announced because of delays providing the environment to our collaborators, Preferred Networks.
However, Preferred Networks worked quickly to provide the baselines in time for participants to use them in Round 1, and many participants found these baselines to be crucial in developing their own algorithms.
In future iterations of this competition, we will be able to provide the baselines as soon as the competition begins now that the baselines are complete.

The data was readily available and participants were generally able to successfully use it.
The provided Gym environments were a wrapper around the Malmo Minecraft environment~\citep{malmo}.
In addition to wrapping Malmo in a standardized way, we fixed bugs and added functionality which allows faster and more efficient training.
These modifications have increased interest in working on Minecraft, and 
we have been contacted by several academic groups about further extensions to the environment.

Competitor submissions had to adhere to a specific Docker format, as outlined in Section~\ref{sec:comp-design}.
Use of a standard container streamlined the evaluation process during both rounds, and it saved time from our end.
Although we provided information about how to use the provided container, based on feedback from participants, we plan to extend the documentation to include a step-by-step demonstration of how to submit an agent using the provided structure.

\subsection{Impact of Evaluation Methodology} \label{subsec:eval_results}
We chose to have two rounds to balance the need for an explicit computational budget with a desire to not limit participation in the competition.
We did not restrict the number of teams who could participate during the Round 1; however, we chose a pre-specified number of teams to move to Round 2.
As a result, we could commit to retraining all submissions during Round 2.
This commitment was possible due to the strict limits on training time as well as the selection of a specific system on which all training and evaluation would be performed.
Due to this structure, we did not use more computational resources than expected for evaluation.

However, these same strict limits on training time led to delays in obtaining final results. 
We delayed the submission deadlines for competitors; as a result, this reduced the time before we were scheduled to announce results. 
Since training had to happen for a fixed number of days on a specific system, we could not parallelize an individual competitor's submission or run it on faster hardware. 
Despite the delay, we will use a similar restriction in future competitions due to the reproducibility and sample efficiency benefits.

As mentioned in Section~\ref{sec:resources}, we provided each team in Round 2 with a compute grant of \$1500USD in Azure credits to enable training and validation of their methods.
Because of the training time restrictions, participants were not incentivized to spend the remainder of their compute time on extending the training time of their final approach.
We believe we allocated a sufficient amount to each team, since some teams had enough compute time remaining to continue to develop their methods even after the competition was over.

Finally, although we effectively utilized a novel visual domain randomization method to ensure generalization and rule compliance, 
competitors spent a great deal of effort shaping the action space using inductive bias. 
To yield fully generalizable competition results in future iterations of this competition,
we plan to remove semantic labels of the environment's action space and apply a random bijective-mapping to the action space during evaluation.

\section{Conclusion}
We ran the MineRL Competition on Sample Efficient Reinforcement Learning Using Human Priors at NeurIPS2019 in order to promote the development of algorithms that: 1) use human demonstrations alongside reinforcement learning, 2) are sample efficient, and 3) are accessible to the general AI community.
We summarized the submissions and looked at the performance of competitors in different rounds.
We discussed the effects of the competition rules, the provided competition materials, and the evaluation methodology.
Based on the largely positive outcome of the competition, we plan to hold the competition again.
Because no team obtained a diamond, we plan to focus on the same task, \texttt{ObtainDiamond}.
We also plan to keep the same paradigm promoting the development of generalizable and robust learning procedures.
When we hold the competition again, we will clarify the rules and provide more demonstrations of how to participate in order to increase participation.

\acks{We thank Microsoft Azure for providing the computational resources for the competition, NVIDIA for providing prizes, AIJ for funding the inclusion grants, and Microsoft Research for funding travel grants.
We also thank AIcrowd for hosting and helping run the competition, and Preferred Networks for providing the baseline method implementations.
Finally, we thank everyone who provided advice or helped organize the competition.}

\newpage
\bibliography{minerl}
\end{document}